\documentclass{article}

\usepackage[main, final]{neurips_2026}

\usepackage[utf8]{inputenc} % allow utf-8 input
\usepackage[T1]{fontenc}    % use 8-bit T1 fonts
\usepackage{hyperref}       % hyperlinks
\usepackage{url}            % simple URL typesetting
\usepackage{booktabs}       % professional-quality tables
\usepackage{amsfonts}       % blackboard math symbols
\usepackage{nicefrac}       % compact symbols for 1/2, etc.
\usepackage{microtype}      % microtypography
\usepackage{xcolor}         % colors
\usepackage{amsmath}
\usepackage{amssymb}
\usepackage{amsfonts}
\usepackage{graphicx}
\usepackage{multirow}
\usepackage{xcolor}
\usepackage{tikz}
\usepackage{pgfplots}
\pgfplotsset{compat=1.18} % 确保版本兼容
\usepackage{booktabs}
\usepackage{caption}
\usepackage{pgfplots}
\usepackage{tikz}
\usepackage{subcaption}
\usepackage{wrapfig}
\usepackage{algorithm}
\usepackage{algorithmic}
\usepackage{enumitem}

\usepackage{bbm}

\title{WorldAct: Activating Monolithic 3D Worlds into Interactive-Ready Object-Centric Scenes}

\author{
  Jichen Hu$^{1}$\thanks{Equal contribution.}, \enspace
  Jiawei Guo$^{1}$\footnotemark[1], \enspace
  Jiazhong Cen$^{1}$\footnotemark[1], \enspace
  Chen Yang$^{2}$, \enspace
  Sikuang Li$^{1}$, \enspace
  Wei Shen$^{1}$\thanks{Corresponding author: Wei Shen, wei.shen@sjtu.edu.cn}
  \\
\textsuperscript{1}MoE Key Lab of Artificial Intelligence, AI Institute, \\
School of Computer Science, Shanghai Jiao Tong University \enspace \textsuperscript{2}Huawei Inc.
}

\begin{document}

\maketitle

\begin{abstract}
Recent 3D world modeling systems based on generative scene synthesis, such as Marble, can create coherent and explorable 3D environments, yet their outputs are typically static monolithic assets with limited editability and physical interaction. This restricts their use in immersive content creation and embodied simulation, where generated worlds must be actively modified and manipulated. To tackle this challenge, we present \textbf{WorldAct}, a framework that converts static generated 3D worlds into editable and interaction-ready scenes. WorldAct uses a multimodal agent to guide scene decomposition, identify actionable objects, reconstruct geometrically aligned object-level meshes for interaction, and restore the residual background via 3D inpainting. The resulting scenes support object-level editing, collision-aware manipulation, and embodied task execution while preserving global scene coherence. Experiments show that WorldAct enables richer interaction scenarios than the original generated scenes, suggesting a practical path toward editable and interactive 3D world models. The project page is at \url{https://sjtu-deepvisionlab.github.io/WorldAct/}
\end{abstract}

\section{Introduction}
% backgroud
Recent advancements in generative modeling have enabled the creation of immersive 3D worlds~\cite{yu2025wonderworld, hy2026hy, schwarz2025recipe, chu2026roamscene3d, hollein2023text2room, chung2023luciddreamer, shriram2024realmdreamer, marble_project} from simple text or image prompts. These models synthesize large-scale, spatially coherent environments, serving as a foundational tool for virtual simulation and digital content creation.

% limitation
Despite these advances, editability and interactivity remain critical limitations. Existing 3D generative world models typically produce static, monolithic 3D representations, where objects are fused into a single structure and cannot be individually selected, moved, or replaced. This limits their use in creative workflows such as game design and interior decoration, where fine-grained scene editing is essential. It also restricts embodied AI simulation, as agents cannot manipulate specific entities in an unstructured scene. Without explicit semantic and physical object decoupling, generated worlds remain inert, serving only as visually plausible environments.

\begin{figure}
    \centering
    \includegraphics[width=\linewidth]{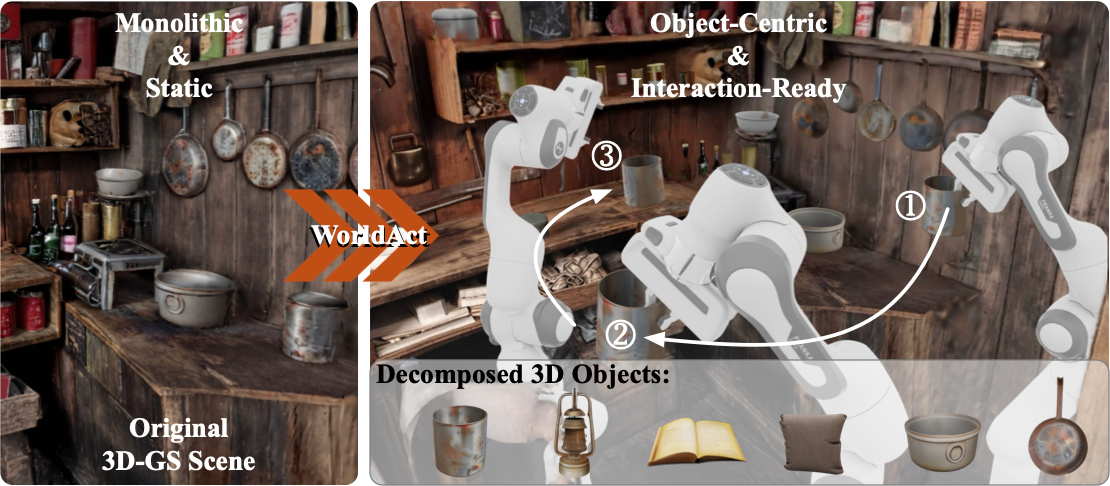}
    \caption{WorldAct converts a monolithic 3DGS scene into a decomposable, object-centric, and interaction-ready environment. By separating individual 3D objects and augmenting them with structures required for physical interaction, our framework enables downstream simulation tasks such as robotic manipulation and scene rearrangement.}
    \label{fig:teaser}
\end{figure}

To address the lack of interactivity in existing 3D generative world models, we present \textbf{WorldAct}, a framework that converts monolithic 3D Gaussian Splatting (3DGS)~\cite{kerbl3Dgaussians} scenes into editable and physically interactive worlds. Given a generated 3DGS scene, WorldAct first uses a vision-language agent to find objects that can be manipulated and select useful viewpoints for scene analysis. The selected views are then segmented in 2D, projected back to 3D, and combined to separate individual objects from the original scene. After removing these objects, WorldAct fills the missing background regions and rebuilds high-quality object assets, which are then placed back into the repaired scene. To support physical interaction, WorldAct also builds simplified collision geometry from the scene, enabling stable placement, collision-aware manipulation, and embodied tasks. In this way, WorldAct turns static monolithic generated worlds into structured scenes where individual objects can be edited, moved, and interacted with.

% contribution
Our key contributions are summarized as follows:
\begin{itemize}
    \item \textbf{Interactive 3D World Modeling.} 
    We propose a framework that converts monolithic 3D generated scenes into decomposed, interaction-ready environments, enabling object-level editing and manipulation.

    \item \textbf{Agent-Driven Automation.} 
    We design an agent-looped pipeline that automatically identifies operable objects, decomposes the scene, restores the background, and reconstructs object assets without manual annotation.

    \item \textbf{Application-Oriented Evaluation.} 
    We evaluate the generated scenes in editing and interaction tasks, demonstrating their visual quality, efficiency, and practical value for downstream applications.
\end{itemize}

\section{Related Works}

% 3d Scene generation
\subsection{3D Scene Generation}

The evolution of 3D representations, from NeRFs~\cite{mildenhall2021nerf} to 3D Gaussian Splatting (3DGS)~\cite{kerbl3Dgaussians}, has enabled efficient and photorealistic rendering of complex scenes. Building on these advances, recent generative methods such as LucidDreamer~\cite{chung2023luciddreamer}, Text2Room~\cite{hollein2023text2room}, Marble~\cite{marble_project}, and HY-World~\cite{hy2026hy} can synthesize complete 3D worlds from text or images. However, these approaches produce static, monolithic representations in which all scene elements are fused together, limiting object-level editing and interaction.

To address this limitation, compositional approaches generate scenes by assembling individual objects. Some methods~\cite{dong2025hiscene,sautter20253d,yao2025cast,wang2025tabletopgen,zhou2025layout} generate objects independently before placing them, while others~\cite{huang2025midi,meng2025scenegen,shi2025scenemaker} jointly model object generation and layout. Agent-based methods~\cite{dai2024acdc,ling2024scenethesis,yang2025sceneweaver,xia2026sage} further leverage asset retrieval for scene construction. While these approaches enable object-level controllability and interaction, they typically rely on limited-view inputs or predefined assets, making it difficult to generate large-scale, multi-view consistent environments with high photorealism.

% Our method bridges these two paradigms by starting from a fully generated 3DGS scene and decomposing it into editable, object-centric components. This preserves the visual fidelity of monolithic generation while enabling object-level interaction and manipulation.
% 3d Scene Decomposition

\subsection{Scene Decomposition and Restoration}

Decomposing a fused 3D scene into individual objects is a key step toward interaction. Recent advances in 2D segmentation, such as SAM~\cite{sam}, and vision-language models, such as CLIP~\cite{radford2021learning}, have inspired a line of methods that lift 2D masks into 3D, including LangSplat~\cite{qin2024langsplat}, Feature3DGS~\cite{zhou2024feature}, and related works~\cite{ye2024gaussian, omniseg3d, sa3d, gaga, saga}. These methods provide useful object-level partitions, but the extracted objects are often incomplete, as they mainly consist of visible Gaussians and lack occluded geometry or clean mesh representations. Meanwhile, removing objects from the scene leaves holes in the background, which can be partially addressed by 3D inpainting methods~\cite{chen2024gaussianeditor, wang2024gaussianeditor, mirzaei2023spin, liu2024infusion, wang2026inpaint360gs, huang20253d}. However, completing large missing regions while preserving scene consistency remains challenging.

\subsection{Object-Level 3D Generation.}
Object-level 3D generation has evolved from SDS-based text-to-3D optimization with frozen 2D diffusion priors~\cite{poole2022dreamfusion,sjc,magic3d,fantasia3d,prolificdreamer,dreamcraft3d,gaussiandreamer,dreamgaussian} to image-conditioned asset generation and reconstruction from single or multi-view inputs~\cite{realfusion,neurallift360,makeit3d,wonder3d,syncdreamer,xu2024instantmesh,li2023instant3d}. Although effective, these 2D-prior-based methods often face limited 3D consistency or expensive optimization. Recent native 3D generative models instead learn directly over 3D representations such as point clouds, voxels, meshes, 3D Gaussians, and neural fields~\cite{nichol2022point,vahdat2022lion,zhang20233dshape2vecset,ren2024xcube,xiong2025octfusion,yang2024atlas}, enabling more efficient geometry generation and textured asset synthesis~\cite{chen2025ultra3d,direct3d,direct3ds2,li2025sparc3d,ye2025hi3dgen,clay,3dtopiaxl,hunyuan3d2025hunyuan3d2.1,lai2025hunyuan3d2.5,step1x,zhou2025few,lin2025partcrafter,unilat}. In particular, SAM3D~\cite{chen2025sam} improves object asset generation under occlusion, making it useful for reconstructing clean objects from complex indoor scenes.

\section{Preliminaries}
\label{sec:preliminaries}

\subsection{3D Gaussian Splatting}
\label{sec:prelim_3dgs}

3D Gaussian Splatting (3DGS)~\cite{kerbl3Dgaussians} represents a continuous 3D scene as an explicit set of unstructured colored Gaussian primitives. 
Let $\mathcal{G}=\{\mathbf{g}_i\}_{i=1}^{N}$ denote a 3DGS scene with $N$ Gaussians, where each primitive is parameterized as
\begin{equation}
    \mathbf{g}_i=(\boldsymbol{\mu}_i,\boldsymbol{\Sigma}_i,\alpha_i,\mathbf{c}_i).
\end{equation}
Here, $\boldsymbol{\mu}_i\in\mathbb{R}^3$ is the 3D center, 
$\boldsymbol{\Sigma}_i\in\mathbb{R}^{3\times 3}$ is the anisotropic covariance, 
$\alpha_i\in[0,1]$ is the opacity, and $\mathbf{c}_i$ denotes the color feature. 
For rendering, Gaussians are projected to the image plane and accumulated by differentiable alpha blending:
\begin{equation}
    \mathbf{C}(\mathbf{p})
    =
    \sum_{i=1}^{K}
    \mathbf{c}_i
    \alpha'_i(\mathbf{p})
    \prod_{j=1}^{i-1}
    \left(1-\alpha'_j(\mathbf{p})\right),
\end{equation}
where $\mathbf{p}$ is a pixel, $K$ is the number of depth-ordered Gaussians overlapping $\mathbf{p}$, and $\alpha'_i(\mathbf{p})$ is the effective opacity of the $i$-th projected Gaussian.

In this work, we mainly consider 3D worlds represented by 3DGS.
Unless otherwise specified, a generated 3D world is denoted as a Gaussian set $\mathcal{G}$, which serves as the renderable visual representation of the scene.

\subsection{3D World Models}
\label{sec:prelim_world_model}

Recent 3D world models aim to generate large-scale, navigable, and spatially coherent 3D environments from sparse conditions such as text, images, videos, or panoramas~\cite{yu2025wonderworld,hy2026hy,schwarz2025recipe,chu2026roamscene3d,hollein2023text2room,chung2023luciddreamer,shriram2024realmdreamer,marble_project}. 
Such models can be abstracted as a conditional generator $\Phi: \mathcal{X}\rightarrow \mathcal{W},$ where $\mathcal{X}$ denotes the input condition and $\mathcal{W}$ denotes the generated 3D world. Generally, the generated world is represented as a 3DGS scene.

Although existing 3D world models have shown impressive visual fidelity, their outputs are still monolithic visual assets rather than interactive-ready environments. Ideally, a 3D world should provide object-level entities, surface or proxy geometry. For a 3DGS-based world, object-level entities can be viewed as a partition of Gaussian primitives:
\begin{equation}
    \mathcal{G}
    =
    \mathcal{G}_{\mathrm{bg}}
    \cup
    \mathcal{G}_{1}
    \cup
    \cdots
    \cup
    \mathcal{G}_{M},
    \quad
    \mathcal{G}_{m}\cap\mathcal{G}_{n}=\varnothing,
\end{equation}
where $\mathcal{G}_{\mathrm{bg}}$ denotes the background and each $\mathcal{G}_{m}$ corresponds to an independently editable object. 
However, standard 3D world models do not directly provide such primitive-to-entity assignments. 
Moreover, raw 3DGS scenes do not explicitly encode watertight surfaces, collision proxies, or physical properties such as mass, friction, and support relations. 
Therefore, despite being visually plausible, existing generated 3D worlds are not directly suitable for downstream tasks like embodied simulation. 

\begin{figure}[!t]
    \centering
    \includegraphics[width=\linewidth]{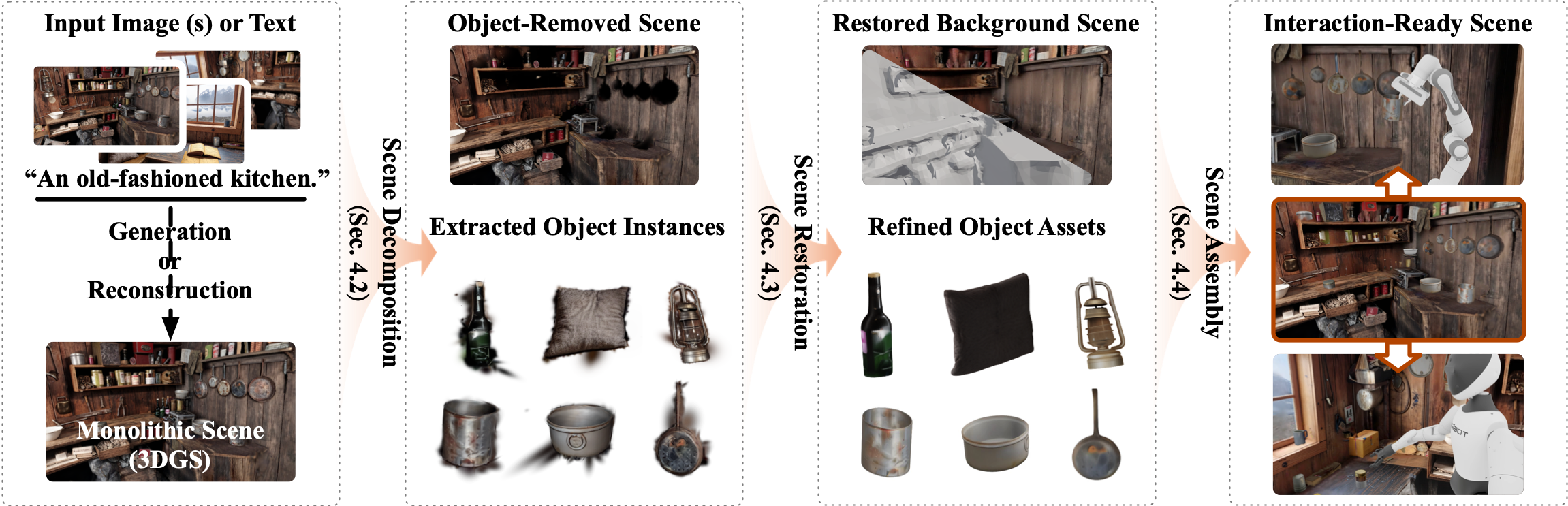}
    \caption{
    WorldAct first decomposes a generated or reconstructed 3DGS scene into an object-removed background and a set of extracted object instances. It then restores the incomplete background, reconstructs scene-level collision geometry, and refines the extracted instances into clean object assets. Finally, WorldAct assembles these assets back into the restored scene, producing an interaction-ready environment with independent object representations.
    }
    \label{fig:pipe}
% \vspace{-5mm}
\end{figure}

\section{Method}
% \vspace{-3mm}
\label{sec:method}
In this section, we first present the overall pipeline of WorldAct, followed by a detailed explanation of each stage.
\subsection{Pipeline Overview}
\label{sec:method_overview}
% pipeline overview

As shown in Figure~\ref{fig:pipe}, WorldAct converts a monolithic 3D Gaussian Splatting (3DGS) scene \(\mathcal{G}\), either generated from text/images or reconstructed from multi-view observations, into an interaction-ready, object-decomposed environment. It first decomposes the scene into an object-removed background and extracted object instances via agent-guided multi-view segmentation and 2D-to-3D mask lifting. The background is then restored with scene-level collision geometry, while the extracted instances are refined into clean object assets. Finally, WorldAct aligns and assembles these assets into a restored scene, where the background and objects are independently represented for editing, manipulation, and embodied interaction.

\subsection{Scene Decomposition}
\label{sec:scene_decomposition}
% Segmentation
Given a monolithic 3DGS scene \(\mathcal{G}\) produced by a 3D world model, WorldAct first renders a camera trajectory to obtain multi-view observations for object discovery and segmentation. Specifically, we define a camera trajectory \(\mathcal{T} = \{\mathbf{T}_t\}_{t=1}^{T}\) that navigates through the scene, capturing a video sequence of RGB frames \(\{\mathbf{I}_t\}_{t=1}^{T}\) along with their camera poses.

\subsubsection{Agent-Driven Interactable Object Discovery}
\label{sec:object_discovery}
\begin{figure}[!t]
    \centering
    \includegraphics[width=\linewidth]{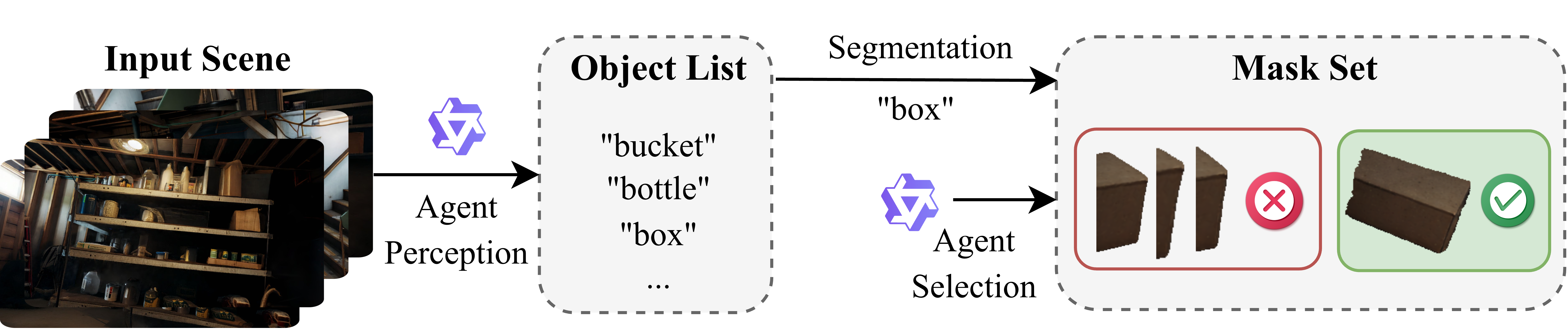}
    \caption{
    Agent-driven object discovery and best frame selection in WorldAct. 
    The agent identifies interactable objects from rendered scene observations and selects the best object view from multi-view masks for reliable 3D asset generation.
    }
    \label{fig:agent}
% \vspace{-5mm}
\end{figure}

To automate object discovery without manual annotation, we employ a vision-language agent (\emph{e.g.}, Qwen3.6-Plus~\cite{qwen36plus}) that analyzes a sparse set of keyframes sampled from the trajectory. As shown in Figure~\ref{fig:agent}, the agent proposes manipulable objects visible in these observations and generates a text prompt list \(\mathcal{P} = \{p_1, \dots, p_N\}\), where each \(p_n\) corresponds to a distinct object prompt such as ``jar'' or ``pillow''. The agent also filters out objects that are semantically irrelevant for interaction.

For each prompt \(p_n\) in \(\mathcal{P}\), we perform video segmentation using SAM3 \cite{carion2025sam}, a promptable segmentation foundation model. We prompt SAM3 with the object's semantic label. The model processes each frame \(\mathbf{I}_t\) to produce a binary mask \(\mathbf{M}_{t,m} \in \{0,1\}^{H \times W}\) indicating the pixel region occupied by the object corresponding to \(p_m\). After processing all prompts, we obtain an object list \(\mathcal{O} = \{o_1, \dots, o_M\}\). The output of this stage is a set of multi-view mask sequences \(\{\mathbf{M}_{t,m}\}_{t=1}^{T}\) for each object \(o_m\), which serve as the input to the subsequent 3D decomposition stage.
\subsubsection{Object-Level 3DGS Segmentation}
\label{sec:object_segmentation}

Given multi-view masks for each object, WorldAct decomposes the input 3DGS scene into object-level Gaussian subsets and a residual background. We denote the input scene as
$
\mathcal{G}=\{g_i\}_{i=1}^{N},
$
where each Gaussian \(g_i\) contains its geometry, opacity, and appearance attributes. For object \(o_m\), we estimate a Gaussian subset
\begin{equation}
\mathcal{G}_m=\{g_i\in\mathcal{G}\mid z_{i,m}=1\},
\end{equation}
where \(z_{i,m}\in\{0,1\}\) indicates whether \(g_i\) belongs to \(o_m\).

Following SA3D~\cite{sa3d}, we propose a learnable soft assignment score \(s_{i,m}\in[0,1]\) for each Gaussian and optimize it through mask inverse rendering. For a view \(t\in\mathcal{V}_m\), let \(\mathbf{M}_{t,m}\in\{0,1\}^{H\times W}\) be the 2D mask of object \(o_m\). The rendered soft mask is computed as
\begin{equation}
\hat{\mathbf{M}}_{t,m}(\mathbf{r})
=
\sum_{i=1}^{N} w_i^t(\mathbf{r})\,s_{i,m},
\end{equation}
where \(\mathbf{r}\) is a pixel ray and \(w_i^t(\mathbf{r})\) is the 3DGS alpha-compositing weight of Gaussian \(g_i\) on this ray. We optimize \(s_{i,m}\) with the projection loss
\begin{equation}
\mathcal{L}_{\mathrm{seg}}^{m}
=
\sum_{t\in\mathcal{V}_m}
\sum_{\mathbf{r}\in\mathcal{R}(\mathbf{I}_t)}
\left[
-\mathbf{M}_{t,m}(\mathbf{r})\hat{\mathbf{M}}_{t,m}(\mathbf{r})
+
\lambda\bigl(1-\mathbf{M}_{t,m}(\mathbf{r})\bigr)\hat{\mathbf{M}}_{t,m}(\mathbf{r})
\right],
\end{equation}
where the first term encourages foreground consistency and the second term suppresses false positives in background regions. During optimization, the 3DGS parameters are fixed and only the assignment scores are updated. After convergence, we binarize the scores by a threshold \(\tau\):
\begin{equation}
z_{i,m} =
\begin{cases}
1, & s_{i,m}>\tau,\\
0, & \text{otherwise},
\end{cases}
\qquad
\mathcal{G}_m=\{g_i\in\mathcal{G}\mid z_{i,m}=1\}.    
\end{equation}
After all objects are processed, the background is defined as
$
    \mathcal{G}_{\mathrm{bg}}
    =
    \mathcal{G}\setminus\bigcup_{m=1}^{M}\mathcal{G}_m .
$
Since \(\mathcal{G}_m\) may still be noisy or incomplete due to occlusion and segmentation errors, we use it only as a spatial proxy for object localization and regenerate clean object assets in the following stage.

\subsection{Scene Restoration}
\label{sec:scene_restoration}

% Inpainting and meshy
\subsubsection{Background Completion}
\label{sec:background_completion}

After removing the object Gaussians, the residual background \(\mathcal{G}_{\mathrm{bg}}\) contains missing regions at the removed object locations. To complete the background, we first build temporally and geometrically consistent removal masks. Given the multi-view object masks \(\{\mathbf{M}_{t,m}\}\), we fuse them into a 3D mask representation through Gaussian splatting reprojection and render it back to each view, obtaining complete masks \(\{\mathbf{M}_{t}^{\mathrm{comp}}\}\) along the trajectory. We then apply Diffueraser~\cite{li2025diffueraser} to the rendered video \(\{\mathbf{I}_t\}\) with the complete masks \(\{\mathbf{M}_{t}^{\mathrm{comp}}\}\), producing inpainted frames \(\{\mathbf{I}_{t}^{\mathrm{inp}}\}\). To lift the inpainted content back to 3D, we select sparse keyframes, estimate their depths using DepthLab~\cite{liu2024depthlab}, and initialize new Gaussians from the predicted depths. Following Infusion~\cite{liu2024infusion}, these Gaussians are then optimized to match the inpainted keyframes, yielding a complete background representation \(\mathcal{G}_{\mathrm{bg}}^{\mathrm{comp}}\).

To enable physical interaction, we further construct a lightweight collision proxy from \(\mathcal{G}_{\mathrm{bg}}^{\mathrm{comp}}\). We extract a watertight mesh using Poisson reconstruction~\cite{kazhdan2006poisson}, then simplify the mesh and regularize major planar structures using plane detection. Specifically, we perform iterative RANSAC to identify planes from uniformly sampled mesh points, classify them by normal orientation (floors/walls/ceilings), and project nearby vertices onto the detected planes to enforce planarity. The resulting low-polygon mesh \(\mathcal{M}_{\mathrm{bg}}\) approximates the background geometry and is used for stable placement and collision-aware simulation.

\subsubsection{Agent-Driven Object Generation}
\label{sec:object_mesh_generation}
% generation
After background repair, we focus on generating high-quality assets for 3D objects. Due to occlusion and incomplete observations in the original scene, the isolated Gaussians \(\mathcal{G}_m\) are often incomplete and not directly usable for interaction. Instead, we adopt SAM3D \cite{chen2025sam}, a feed-forward model that generates complete 3DGS and mesh assets from single-view RGB images and masks. However, not all viewpoints are equally suitable for generation, as occlusion or unfavorable angles can degrade the output. To address this, we employ an agent to automatically select the optimal viewpoint for each object by evaluating visibility, occlusion levels, and semantic confidence across all frames in the trajectory, as illustrated in Figure~\ref{fig:agent}. The agent then feeds the selected RGB image and its corresponding mask into SAM3D, which produces a clean 3DGS representation \(\mathcal{G}_m^{\text{gen}}\) and a textured mesh \(\mathcal{M}_m^{\text{gen}}\) for object \(o_m\).

\subsection{Scene Assembly}
\label{sec:scene_assembly}

Although SAM3D provides an estimated pose for each generated object, we observe that the predicted pose is often inaccurate and may not align well with the restored scene. To place each generated object into the completed background \(\mathcal{G}_{\mathrm{bg}}^{\mathrm{comp}}\), we use a two-stage alignment procedure.

First, we estimate an initial pose using the extracted object Gaussians \(\mathcal{G}_m\) as spatial anchors. Given the generated object mesh \(\mathcal{M}_{m}^{\mathrm{gen}}\), we perform Iterative Closest Point (ICP) between \(\mathcal{M}_{m}^{\mathrm{gen}}\) and the point set derived from \(\mathcal{G}_m\) under multiple candidate transformations. For each candidate pose, we render the placed object and compare it with the original object observations using DINOv2~\cite{oquab2023dinov2} features. The pose with the highest feature similarity is selected as the initialization.

Second, we refine the object pose through differentiable rendering. For each object \(o_m\), we optimize its translation \(\mathbf{t}_m\in\mathbb{R}^3\), rotation represented in 6D form \(\mathbf{r}_m\in\mathbb{R}^6\), and scale \(s_m\in\mathbb{R}^{+}\). The optimization minimizes
\begin{equation}
\mathcal{L}_{\mathrm{align}}
=
\mathcal{L}_{\mathrm{mask}}
+
w_c \mathcal{L}_{\mathrm{contact}}
+
w_p \mathcal{L}_{\mathrm{penetration}},
\end{equation}
Here, \(\mathcal{L}_{\mathrm{mask}}\) combines soft Dice and
focal losses between the rendered object silhouettes and
the original object masks across sampled views. \(\mathcal{L}_{\mathrm{contact}}\) applies a smooth L1 penalty
to small positive vertical gaps between the object and nearby background surfaces,
encouraging support contact. \(\mathcal{L}_{\mathrm{penetration}}\) identifies penetrating
object vertices using nearest background surface samples
and their oriented normals, and penalizes the distances to
those samples. We use \(w_c=0.01\) and \(w_p=0.02\).

After alignment, the final scene consists of the completed background \(\mathcal{G}_{\mathrm{bg}}^{\mathrm{comp}}\) with its collision mesh \(\mathcal{M}_{\mathrm{bg}}\), together with a set of generated object assets \(\{(\mathcal{G}_{m}^{\mathrm{gen}}, \mathcal{M}_{m}^{\mathrm{gen}})\}_{m=1}^{M}\) placed in the scene. This decomposed representation supports object-level editing, manipulation, and embodied task execution.

\begin{figure}[!t]
    \centering
    \includegraphics[width=\linewidth]{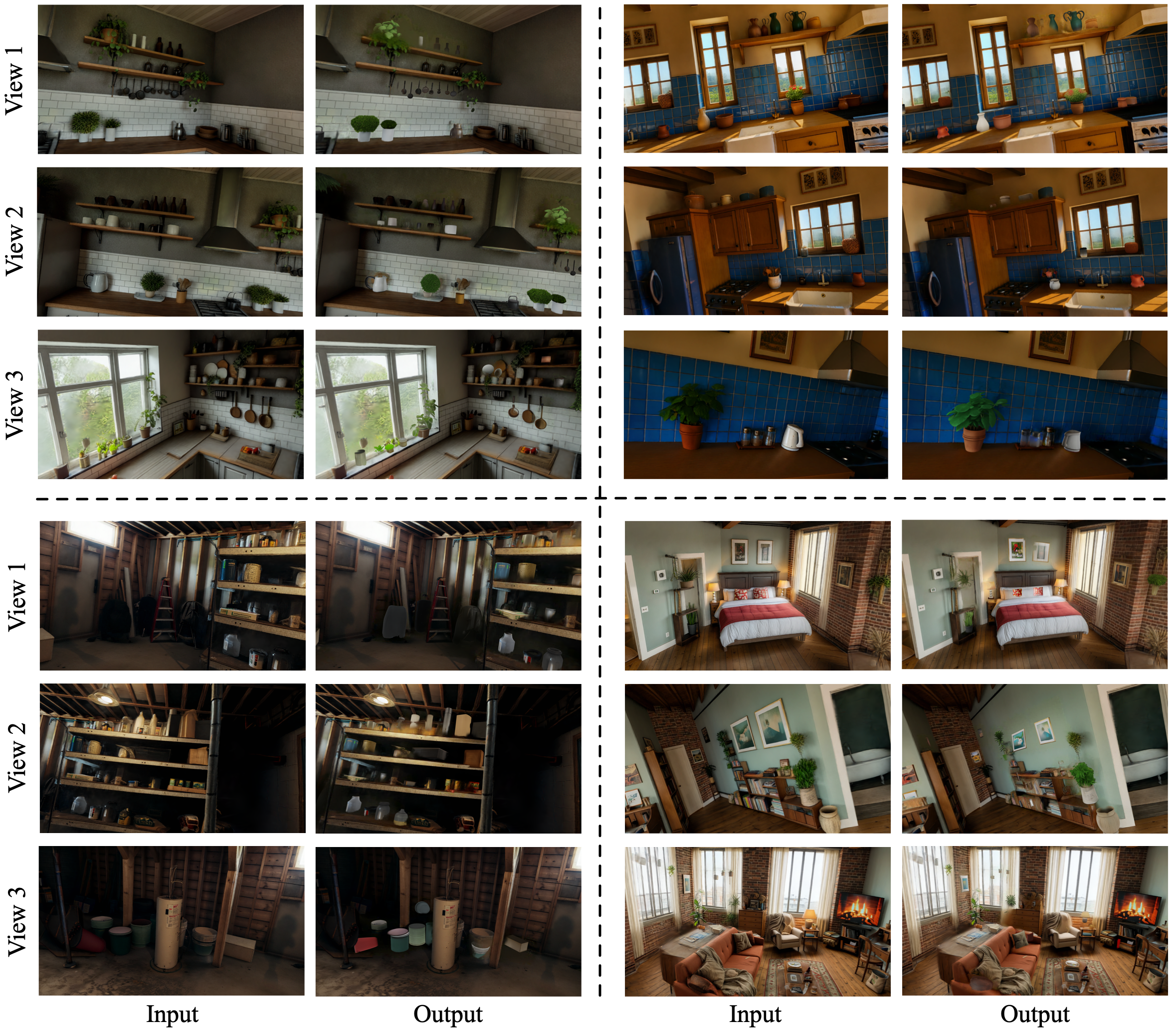}
    \caption{Qualitative comparison with input scenes. For each scene, we show three different viewpoints, each with the original input and our decomposed interactive output. Our method preserves visual fidelity while enabling object-level decomposition and interaction.}
    \label{fig:visualization}
\end{figure}

\section{Experiments}
\subsection{Implementation Details}

All experiments are conducted on a single NVIDIA RTX 3090 GPU. Converting a 3DGS scene typically takes around 1 hour, varying with scene complexity.

% We evaluate our framework on six diverse indoor scenes generated by Marble~\cite{marble_project}. These scenes cover different architectural styles, including functional categories such as kitchen, restroom and storage room. We choose Marble as our primary foundation model not only for its strong generation quality, but also because it represents a typical 3D world model: it can take text, single-image, or multi-image inputs and produce monolithic 3DGS scenes. This makes it a suitable testbed for studying whether generated 3D worlds can be further decomposed, repaired, and converted into interaction-ready environments. To assess performance across different foundation models, we additionally evaluate on four scenes generated by HY-World 2.0~\cite{hy2026hy}.
We evaluate on six indoor scenes generated by Marble~\cite{marble_project}, including kitchens, restrooms, and storage rooms. Marble accepts text, single-image, or multi-image inputs and produces monolithic 3DGS scenes. We additionally evaluate on four scenes generated by HY-World 2.0~\cite{hy2026hy} to assess performance across foundation models.

As ground-truth decomposed assets and completed backgrounds
are unavailable, we primarily use reference-free metrics to
assess conversion quality. ReMOVE~\cite{chandrasekar2024remove,
zhao2025objectclear} measures foreground-background consistency
after object removal, while MANIQA~\cite{yang2022maniqa}
evaluates perceptual image quality. For decomposition, we
report \textit{Interactable Object Recall}, the fraction of
manually annotated interactable objects that are successfully
extracted. We further evaluate scene and object quality
through a Mean Opinion Score (MOS) study with 20 participants
using a 5-point Likert scale across four dimensions: overall
quality, surface completeness, boundary cleanliness, and
naturalness. As an auxiliary evaluator,
GPT-5.5~\cite{openai2026systemcard} rates matched scene and
object renderings from the original and converted Marble
scenes using the same criteria. As a supplementary
reference-based comparison, we report PSNR, SSIM, and LPIPS
on selected keyframes, using the original 3DGS renderings
from the same viewpoints as references.

\begin{figure}[!t]
    \centering
    \includegraphics[width=1.0\linewidth]{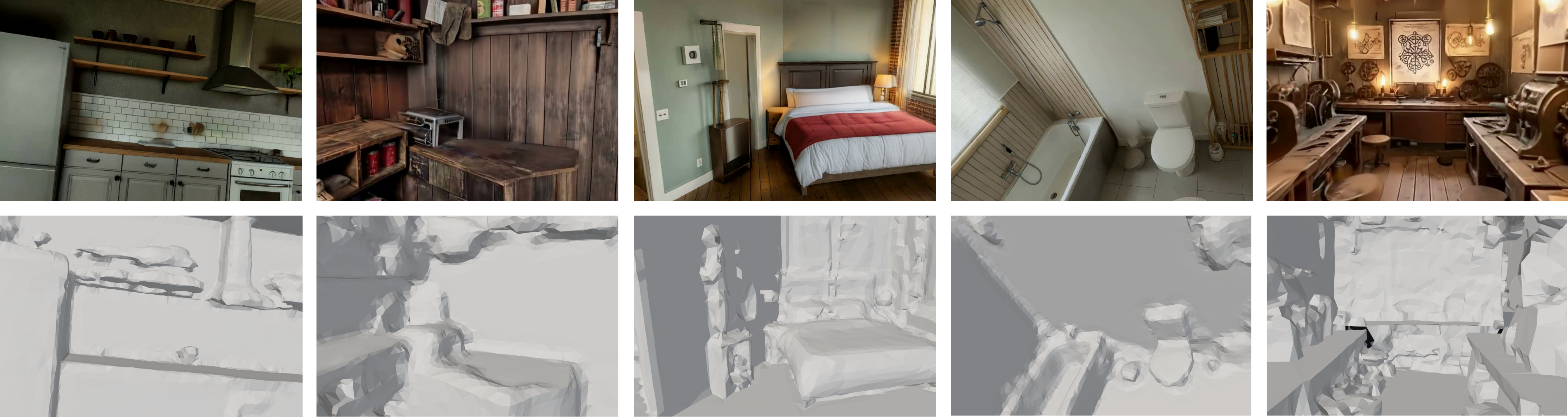}
    \caption{
    Qualitative results of background mesh reconstruction.
    Top: background renderings after object removal and inpainting.
    Bottom: the corresponding reconstructed meshes used as background collision geometry.
    }
    \label{fig:background_mesh}
\end{figure}

\textbf{Qualitative Results.}
Figure~\ref{fig:visualization} compares the input scenes with the converted scenes from multiple viewpoints. After decomposition and regeneration, objects are placed back into their original locations within the restored background. The results retain the overall scene layout and show coherent object boundaries across views. At the object level, regeneration completes some missing surfaces and reduces visible deformations and blurriness, while allowing the resulting assets to be edited independently. Figure~\ref{fig:background_mesh} shows the extracted meshes alongside the inpainted backgrounds. The meshes align well with the restored scene geometry and exhibit relatively smooth surfaces, providing collision geometry for downstream simulation.

\begin{table}[!t]
    \centering
    \caption{\textbf{Interactable Object Recall Rate.} Object discovery recall on Marble and HY-World. Higher is better.}
    \label{tab:recall_rate}
    \begin{tabular}{lcc}
    \toprule
    Method & Marble (\%) $\uparrow$ & HY-World (\%) $\uparrow$ \\
    \midrule
    Ours w/o Agent & 23.77 &  8.10\\
    Ours & \textbf{78.80} & \textbf{70.97} \\
    \bottomrule
    \end{tabular}
\end{table}

\textbf{Quantitative Results.}
% Table~\ref{tab:recall_rate} reports the Interactable Object Recall Rate. On MWM, agent guidance increases recall from 23.77\% to 78.80\%, showing its effectiveness in identifying interactable objects. Objects missed during discovery remain in the background and are not converted into independently editable assets.
Table~\ref{tab:recall_rate} reports the Interactable Object Recall Rate. Agent guidance increases recall from 23.77\% to 78.80\% on Marble and from 8.10\% to 70.97\% on HY-World, showing its effectiveness in identifying interactable objects across both datasets. Objects missed during discovery remain in the background and are not converted into independently editable assets.

\begin{table}[!t]
    \centering
    \caption{\textbf{Perceptual quality.} ReMOVE~\cite{chandrasekar2024remove,zhao2025objectclear} and MANIQA~\cite{yang2022maniqa} at different conversion stages. Higher is better.}
    \label{tab:remove_metric}
    \begingroup
    \small
    \setlength{\tabcolsep}{5pt}
    \begin{tabular}{@{}lrrrr@{}}
    \toprule
    & \multicolumn{2}{c}{Marble (6)} & \multicolumn{2}{c}{HY-World (4)} \\
    \cmidrule(lr){2-3}\cmidrule(lr){4-5}
    Stage & ReMOVE & MANIQA & ReMOVE & MANIQA \\
    \midrule
    Original 3DGS & 0.7933 & 0.3338 & 0.7393 & 0.3924 \\
    Object re-insertion & \textbf{0.7934} & \textbf{0.3622} & \textbf{0.7501} & \textbf{0.3951} \\
    \midrule
    Inpainting (Gaussian Grouping) & 0.7948 & 0.3227 & 0.7567 & 0.3729 \\
    Inpainting (ours) & \textbf{0.7951} & \textbf{0.3254} & \textbf{0.7583} & \textbf{0.3740} \\
    \bottomrule
    \end{tabular}
    \endgroup
\end{table}

\begin{table}[!t]
    \centering
    \caption{\textbf{Reference-based comparison.}
    Results on six Marble and four HY-World scenes,
    using one keyframe per scene.}
    \label{tab:reference_fidelity}
    \begingroup
    \setlength{\tabcolsep}{5pt}
    \begin{tabular}{@{}lrrrrrr@{}}
    \toprule
    & \multicolumn{3}{c}{Marble (6)}
    & \multicolumn{3}{c}{HY-World (4)} \\
    \cmidrule(lr){2-4}\cmidrule(lr){5-7}
    Method
    & PSNR $\uparrow$ & SSIM $\uparrow$ & LPIPS $\downarrow$
    & PSNR $\uparrow$ & SSIM $\uparrow$ & LPIPS $\downarrow$ \\
    \midrule
    SAM3D & 24.33 & 0.890 & 0.107 & 25.18 & 0.920 & 0.060 \\
    MIDI  & 21.72 & 0.841 & 0.179 & 23.90 & 0.920 & 0.078 \\
    Ours  & \textbf{25.26} & \textbf{0.903} & \textbf{0.089}
          & \textbf{25.53} & \textbf{0.925} & \textbf{0.058} \\
    \bottomrule
    \end{tabular}
    \endgroup
\end{table}

% CR-TODO T06: Verify evaluation resolution, image region, compositing/occlusion,
% baseline scale and pose alignment, object selection, and aggregation rules.

% Table~\ref{tab:remove_metric} further evaluates perceptual quality using ReMOVE and MANIQA. After object re-insertion, MANIQA increases from 0.3338 to 0.3621 on Marble scenes and from 0.3924 to 0.3951 on HY-World scenes. ReMOVE remains nearly unchanged on Marble and increases from 0.7393 to 0.7501 on HY-World. We also compare background inpainting with Gaussian Grouping on Marble scenes using the same removal masks. Our method obtains slightly higher scores on both metrics. These results complement the reference-based comparison by assessing perceptual quality at different stages of the pipeline.
Table~\ref{tab:remove_metric} summarizes perceptual quality at different conversion stages on Marble and HY-World. Object re-insertion improves MANIQA on both datasets. We also include Gaussian Grouping as a reference for
evaluating background inpainting.

% Table~\ref{tab:remove_metric} evaluates perceptual quality on Marble scenes using ReMOVE and MANIQA. After object re-insertion, MANIQA increases from 0.3338 to 0.3621, while ReMOVE remains nearly unchanged. For background inpainting, our method obtains slightly higher scores than Gaussian Grouping under the same removal masks.
% \begin{table}[t]
%     \centering
%     \caption{\textbf{Perceptual Metric Evaluation.} ReMOVE~\cite{chandrasekar2024remove,zhao2025objectclear} and MANIQA~\cite{yang2022maniqa} at different stages of scene conversion. Higher is better. Inpainting methods use the same removal masks.}
%     \label{tab:remove_metric}
%     \begin{tabular}{lcc}
%     \toprule
%     Stage & ReMOVE $\uparrow$ & MANIQA $\uparrow$ \\
%     \midrule
%     Original 3DGS & 0.7933 & 0.3338 \\
%     After Object Re-insertion (ours) & \textbf{0.7934} & \textbf{0.3621} \\
%     \midrule
%     After Inpainting (Gaussian Grouping) & 0.7948 & 0.3226 \\
%     After Inpainting (ours) & \textbf{0.7951} & \textbf{0.3254} \\
%     \bottomrule
%     \end{tabular}
% \end{table}

WorldAct converts existing 3D scenes into interactive
environments, a task that differs from 3D generation and
makes direct comparison challenging. Nevertheless, we
provide a supplementary reference-based comparison with
SAM3D~\cite{chen2025sam} and MIDI~\cite{huang2025midi} on keyframes from six Marble and four HY-World
scenes. Baseline object renderings are composited with the original background. Table~\ref{tab:reference_fidelity}
reports PSNR, SSIM, and LPIPS computed against the input renderings from the same viewpoints.

\subsection{Interactive Experiments}

\begin{figure}[!t]
    \centering
    \includegraphics[width=1.0\linewidth]{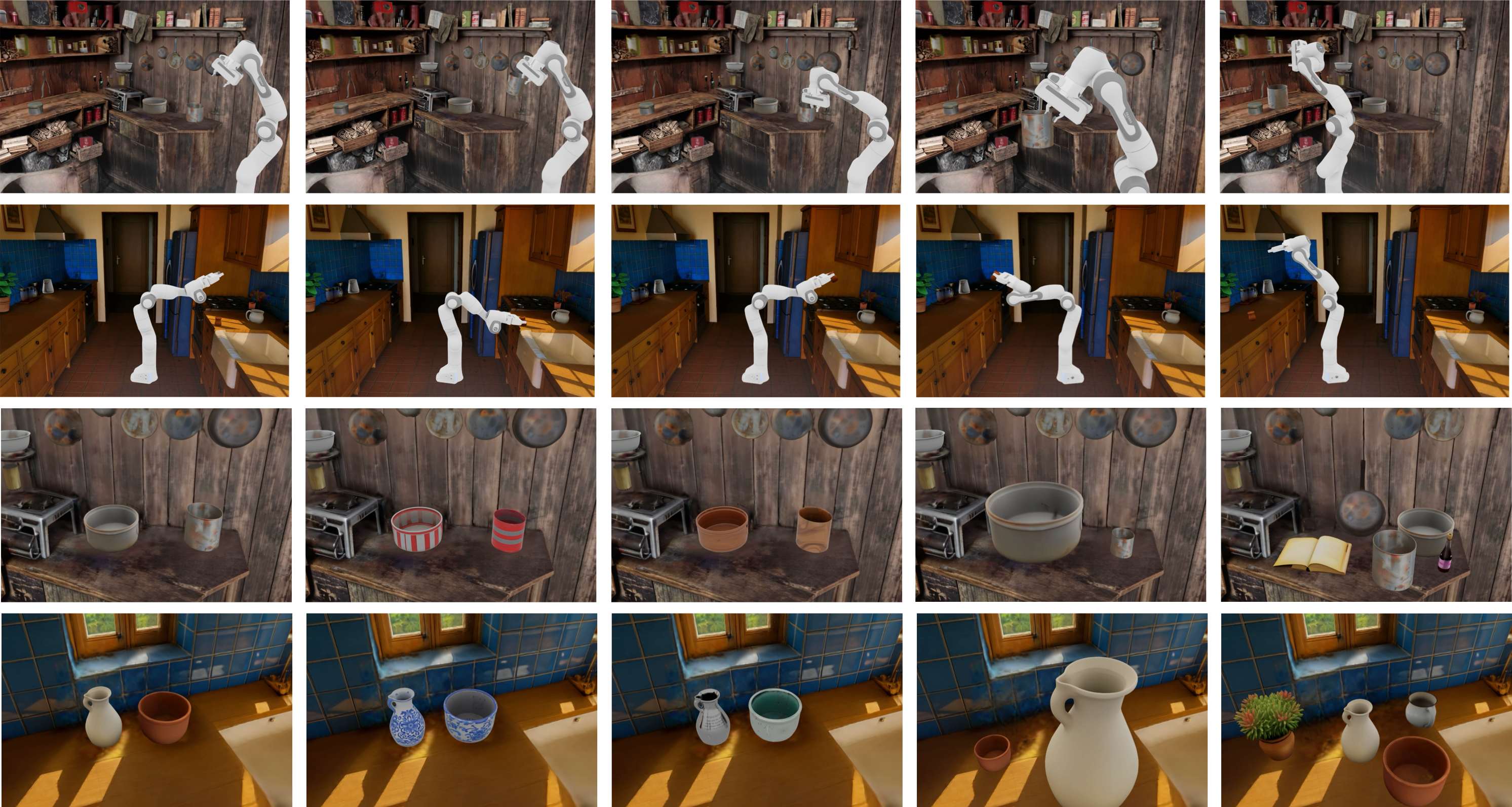}
    \caption{
    Interactive examples with WorldAct. By decomposing a generated 3DGS scene into editable object assets and a restored background, WorldAct enables object-level interaction in 3D worlds. Users can add, place, remove, and modify objects, including size, texture, and material. These capabilities support embodied simulation, scene rearrangement, and interactive content creation.
    }
    \label{fig:interactive example}
\end{figure}

\begin{figure}[!t]
    \centering
    \includegraphics[width=\textwidth]{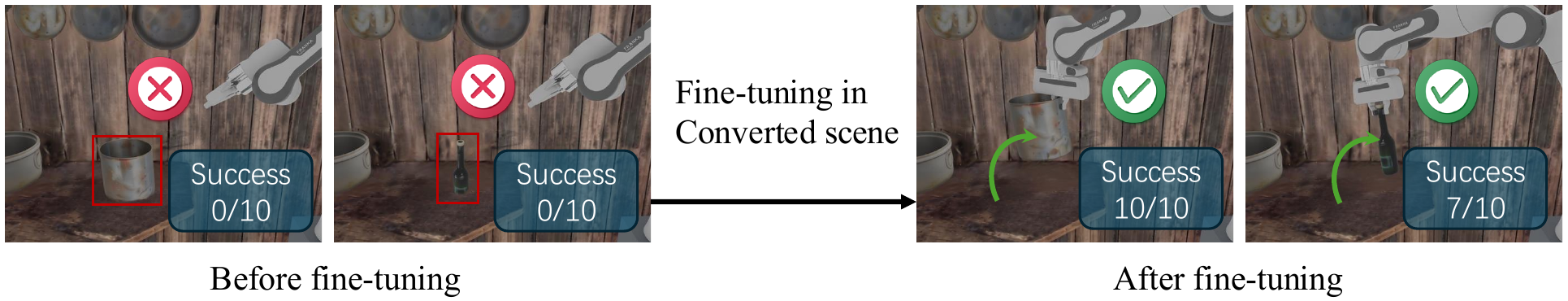}
    \caption{
    Grasping before and after policy fine-tuning in our converted scene. The pretrained policy fails to grasp the cup and thin bottle, while the fine-tuned policies successfully grasp and lift them.
    }
    \label{fig:sim_finetuning}
\end{figure}

\textbf{Application to Embodied Simulation.}
% WorldAct converts a generated 3D scene into an interaction-ready environment for embodied simulation. As shown in the first and second rows of Figure~\ref{fig:interactive example}, after decomposing the scene into explicit object assets and a restored background, our framework supports object-level physical interaction, like grasping and placement. A robotic manipulator can actively interact with objects in the reconstructed kitchen scene, demonstrating that the resulting representation is not limited to passive rendering but can serve as an executable environment for agent--scene interaction. This is difficult for conventional monolithic 3D world representations, where object semantics, geometry, and appearance are tightly entangled. By exposing objects as manipulable entities while preserving coherent scene layout and appearance, WorldAct provides a practical basis for downstream embodied AI tasks such as rearrangement, task planning, and closed-loop simulation.
WorldAct converts a generated 3D scene into an interaction-ready environment for embodied simulation. As shown in the first and second rows of Figure~\ref{fig:interactive example}, after decomposing the scene into explicit object assets and a restored background, our framework supports object-level physical interaction, like grasping and placement. We evaluate language-conditioned grasping in our constructed Isaac Sim environment using a Franka arm and a SmolVLA policy~\cite{shukor2025smolvla} initialized from a LIBERO-trained checkpoint~\cite{liu2023libero}. Given instructions such as ``Pick up the cup,'' the pretrained policy achieves 0/10 successful grasps for both the cup and the thin bottle. For each object, a scripted controller automatically collects 24 successful expert trajectories, split into 20 for training and 4 for validation. We then fine-tune a separate policy for each object for 4,000 steps. Evaluated under the same initial conditions as the pretrained policy, the fine-tuned policies achieve 10/10 successful grasps for the cup and 7/10 for the thin bottle (Fig.~\ref{fig:sim_finetuning}). These results illustrate how the constructed environment supports automated demonstration collection and adaptation of pretrained manipulation policies.

\textbf{Application to High-Quality 3D Scene Editing.}
% WorldAct also supports high-quality object-level scene editing and reconstruction. As shown in the third and fourth rows of Figure~\ref{fig:interactive example}, users can add external assets, remove existing objects, move objects to new locations, and modify their scale, texture, or material properties while maintaining visual coherence with the surrounding scene. In particular, the removal examples show that previously occluded regions can be restored cleanly without obvious holes or object-shaped artifacts, while insertion and attribute editing preserve plausible spatial alignment and appearance consistency. These results demonstrate that WorldAct transforms a globally entangled generated scene into locally editable object assets and a coherent reconstructed background, making 3D world models more suitable for interactive content creation, scene refinement, and controllable reconstruction.
WorldAct supports adding, removing, and relocating objects, as well as editing their scale, texture, and material (Figure~\ref{fig:interactive example}, third and fourth rows). The examples show restored backgrounds after removal and coherent placement of inserted or modified assets.

\subsection{User Study}
\label{sec:user_study}

We conduct a Mean Opinion Score (MOS) study with 20 participants on the six Marble scenes. Participants rate anonymized scene and object renderings on a five-point scale across four criteria: Overall Quality, Surface Completeness, Boundary Cleanliness, and Naturalness. We additionally use GPT-5.5 to evaluate 50 matched scene pairs and 50 matched object pairs from the same scenes, using the same criteria and scale. Human ratings serve as the primary subjective evaluation, with GPT ratings providing an auxiliary assessment. The evaluation protocols and GPT prompt are provided in Appendix~\ref{sec:supp_user_study}.

Table~\ref{tab:user_study} reports both human and GPT ratings. Both evaluations show higher scores for the regenerated objects across all four criteria, alongside lower scores for the converted scenes. For example, object surface completeness increases from 2.58 to 3.93 in the human study and from 2.24 to 3.20 in the GPT evaluation. Scene naturalness decreases from 4.20 to 3.78 and from 3.06 to 2.78, respectively. These results indicate that object regeneration improves the quality of individual assets, while the conversion introduces some loss in scene-level perceptual quality.

\begin{table}[t]
    \centering
    \caption{\textbf{Human and GPT-5.5 ratings.} Each cell reports \textit{Human / GPT} scores on a 1--5 scale, where higher is better. The human study includes 20 participants; the GPT evaluation uses 50 matched pairs per level. Both evaluations cover the same six Marble scenes and are aggregated separately.}
    \label{tab:user_study}
    \setlength{\tabcolsep}{4pt}
    \begin{tabular}{lcccc}
    \toprule
    Evaluated Target 
    & Overall Quality 
    & Surface Comp. 
    & Boundary Clean. 
    & Naturalness \\
    \midrule
    \multicolumn{5}{c}{\textit{Scene-Level Evaluation}} \\
    \midrule
    Original Scene & 4.41 / 3.04 & 4.44 / 3.62 & 4.28 / 2.90 & 4.20 / 3.06 \\
    Converted Scene & 4.13 / 2.78 & 4.31 / 3.26 & 4.02 / 2.60 & 3.78 / 2.78 \\
    \midrule
    \multicolumn{5}{c}{\textit{Object-Level Evaluation}} \\
    \midrule
    Original Object & 2.49 / 1.78 & 2.58 / 2.24 & 2.00 / 1.46 & 2.14 / 1.88 \\
    Regenerated Object & 3.75 / 2.50 & 3.93 / 3.20 & 3.95 / 2.42 & 3.37 / 2.62 \\
    \bottomrule
    \end{tabular}
\end{table}

% \begin{figure*}[t]
%     \centering
%     % \includegraphics[width=\textwidth]{figures/agent_reasoning.pdf}
%     \vspace{4.5cm} % Placeholder height
%     \caption{\textbf{Agent Evaluation Examples.} We showcase the VLM Agent's (GPT-5.3) reasoning process. \textbf{Top:} Scene-level comparison where the agent evaluates the \textit{Naturalness} of the inpainted background after an object is moved, providing text-based justification for its score. \textbf{Bottom:} Object-level comparison detailing the agent's assessment of \textit{Boundary Cleanliness} and \textit{Surface Completeness} for the extracted mesh versus the original point cloud cluster.}
%     \label{fig:agent_reasoning}
% \end{figure*}

\section{Conclusion}
In this paper, we present WorldAct, a framework that converts monolithic 3DGS scenes into object-decomposed environments for editing and interaction. WorldAct identifies objects in a generated scene, separates them from the background, repairs the remaining scene, regenerates cleaner object assets, and aligns them with simple collision geometry. Our experiments show that the pipeline preserves the visual appearance of the original scene while enabling basic object-level editing, placement, and embodied interaction.

% \textbf{Limitations.} The current framework depends on the quality of the input 3D world model and does not yet handle dynamic scenes, articulated objects, or physical properties such as mass and friction. Addressing these limitations is an important direction for future work.
\textbf{Limitations.}
Physical parameters such as mass and friction are currently specified manually, and interactions are limited to rigid objects, without support for articulated objects or more complex dynamics. The pipeline also requires multiple processing stages. Future work will focus on estimating physical properties, extending interaction capabilities, and integrating agent reasoning more closely with 3D generation, while streamlining the pipeline to construct interactive environments more efficiently.

% \section{Acknowledgments}
\begin{ack}
This work was supported by the NSFC under Grants 62322604 and 62576207. The authors
express gratitude to the anonymous reviewers for their valuable feedback. 
\end{ack}

\newpage
\bibliographystyle{plain}
\bibliography{main}

%%%%%%%%%%%%%%%%%%%%%%%%%%%%%%%%%%%%%%%%%%%%%%%%%%%%%%%%%%%%
\newpage
\appendix

\section{Overview}
This supplementary material provides additional details and analyses to complement the main manuscript. Specifically, Section~\ref{app:agent_impl} presents the detailed design of the agent utilized in Sections~\ref{sec:scene_decomposition} and~\ref{sec:scene_restoration}. Furthermore, we elaborate on the design and implementation of the user study in Section~\ref{sec:supp_user_study}. Section~\ref{app:more_results} then showcases additional experimental results of our proposed reconstruction method.

\section{Agent Design Details}
\label{app:agent_impl}

This section details the implementation of our object- and view-selection agent. The agent automatically identifies portable objects from multi-view observations, extracts object masks via text-guided segmentation, and provides the resulting masks and inpainted videos for downstream reconstruction. The agent comprises three core components: a vision module, a memory module, and an execution module.

\subsection{Overview}
\label{app:agent_overview}

Given visual observations $\mathcal{O}=\{I_v\}_{v=1}^{V}$ or a video $\mathcal{V}$, the agent first parses the scene into a structured object inventory. It selects portable objects, generates text prompts for segmentation, invokes SAM3 to extract candidate masks across rendered views, and employs a Vision-Language Model (VLM) to score and select the optimal view for each object. Following mask aggregation and duplicate removal, the agent performs video inpainting to visually erase the selected objects. Formally, the agent outputs a set of object-level representations:
\begin{equation}
    \mathcal{Y} = \{(q_i, v_i^\star, M_i, \mathcal{V}_i^{\mathrm{rm}})\}_{i=1}^{N},
\end{equation}
where $q_i$ is the text prompt of the $i$-th selected object, $v_i^\star$ is its best view, $M_i$ is the aggregated mask, and $\mathcal{V}_i^{\mathrm{rm}}$ is the inpainted background video. These outputs isolate portable objects from the static scene, facilitating the downstream modeling of interactable 3D environments.

\subsection{Vision Module}
\label{app:vision_module}

The vision module converts raw visual inputs into structured semantic descriptions via the Qwen3.6-Plus API. The VLM is prompted to enumerate all visible objects and classify them along two orthogonal axes: mobility (\texttt{portable} vs.\ \texttt{fixed}) and semantic recognizability (\texttt{precise}, \texttt{subtle}, or \texttt{unrecognizable}). This process produces a comprehensive, deduplicated object inventory that systematically accounts for items with ambiguous boundaries or semantics.

\subsection{Memory Module}
\label{app:memory_module}

The memory module maintains the state of the parsed object inventory. For the $i$-th discovered object, it stores a state dictionary:
\begin{equation}
    \mathcal{M}_i = \{ \texttt{name}_i, \texttt{category}_i, \texttt{count}_i, \texttt{recognizability}_i \}.
\end{equation}
Serving as the agent's central representation, the memory module supplies the execution module with exact object-level prompts and tracks processing states to guarantee logical consistency across the pipeline.

\subsection{Execution Module}
\label{app:execution_module}

The execution module handles segmentation, view selection, and video inpainting. For each object categorized as \texttt{portable} in memory, its \texttt{name} serves as a text prompt $q_i$ for SAM3 to generate candidate masks across all available views.

To resolve viewpoint variations, occlusions, and scale inconsistencies, we propose a VLM-based view-scoring mechanism. We prompt Qwen3.6-Plus to evaluate each candidate crop, returning an integer score from 0 to 100 based on the completeness, clarity, and centeredness of the target object. Given the score $s_{i,v}$ for object $i$ in view $v$, the optimal view is selected as:
\begin{equation}
    v_i^\star = \arg\max_{v} s_{i,v}.
\end{equation}
Candidate masks for the same object are subsequently aggregated. After filtering out duplicate masks to prevent redundant selection, the final mask set guides the video inpainting process, yielding a clean background video.

\subsection{Agent Workflow and Interface to Reconstruction}
\label{app:agent_workflow}

The complete agent pipeline is summarized in Alg.~\ref{alg:agent_workflow}. The final outputs—best-view masks and inpainted background videos—provide explicit object-level supervision. By cleanly separating portable objects from the static environment prior to reconstruction, our approach significantly reduces the entanglement between object geometry, appearance, and the background, directly supporting the downstream decomposition of interactive 3D scenes.

\begin{algorithm}[ht]
\caption{Object-Selection Agent Workflow}
\label{alg:agent_workflow}
\begin{algorithmic}[1]
\REQUIRE Input video $\mathcal{V}$ or rendered views $\mathcal{O}$; SAM3 threshold $\tau$
\ENSURE Object masks $M$, best views $v^\star$, and inpainted videos $\mathcal{V}^{\mathrm{rm}}$
\STATE Parse $\mathcal{V}$ or $\mathcal{O}$ into a structured object inventory via the vision module
\STATE Initialize memory module $\mathcal{M}$ with object properties (\texttt{name}, \texttt{category}, etc.)
\FOR{each \texttt{portable} object $i$ in $\mathcal{M}$}
    \STATE Generate candidate masks across views using SAM3 with prompt $q_i = \texttt{name}_i$
    \STATE Score candidate crops via VLM to evaluate target completeness and clarity
    \STATE Select the best view $v_i^\star = \arg\max_v s_{i,v}$
    \STATE Aggregate candidate masks corresponding to object $i$
\ENDFOR
\STATE Filter duplicate object masks
\STATE Generate inpainted video $\mathcal{V}^{\mathrm{rm}}$ using the aggregated masks
\STATE \textbf{return} $\{(q_i, v_i^\star, M_i, \mathcal{V}_i^{\mathrm{rm}})\}_{i=1}^{N}$
\end{algorithmic}
\end{algorithm}

\section{User Study and Auxiliary Agent Evaluation}
\label{sec:supp_user_study}
\begin{figure}[t]
    \centering
    \includegraphics[width=\linewidth]{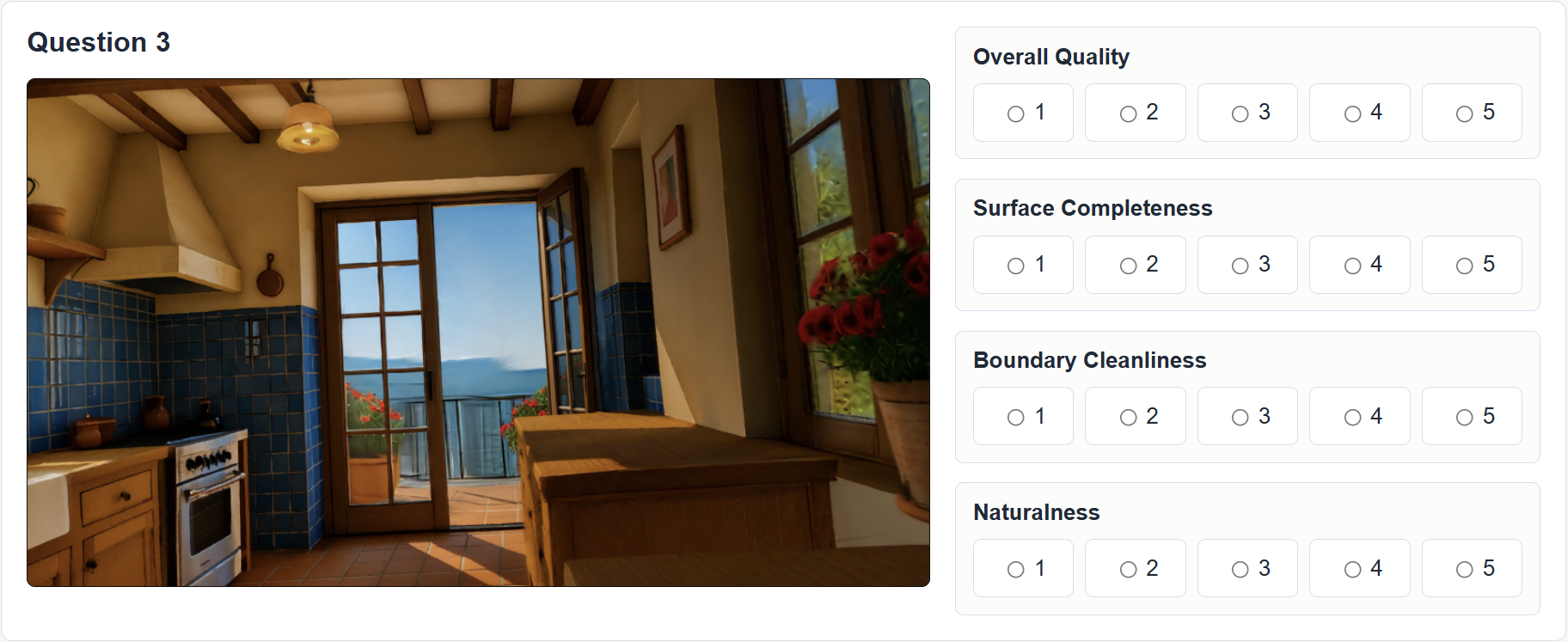}
    
    \caption{Screenshot of the web-based user-study interface. Participants first read the bilingual evaluation criteria and MOS-style score definitions, then rate one anonymized image at a time.}
    \label{fig:supp_user_study_interface}
\end{figure}

\subsection{Human User Study}
\label{sec:supp_human_user_study}

We conducted a blind user study to evaluate the perceptual quality of the rendered results produced by Marble and our method. The study was implemented as a web-based questionnaire. Each questionnaire contained 20 single-image questions, consisting of 10 whole-scene renderings and 10 object-level renderings. We distributed 20 questionnaires in total.

We evaluate all the scene in the Marble dataset. For each question, the participant saw only one anonymized rendering. The questionnaire did not reveal the scene name, object identity, frame index, or method label. These metadata were stored only on the server for later aggregation.

To reduce recognition bias and direct pairwise comparison effects, each questionnaire was generated using randomized sampling under two constraints. First, within a single questionnaire, each scene was assigned to only one method, either Marble or ours. Therefore, a participant never saw both methods for the same scene in the same questionnaire. Second, whole-scene images sampled from the same scene were required to be temporally separated by at least 10 frames. For object-level questions, each object identity appeared at most once in a questionnaire and was shown using only one method.

Each image was rated along four perceptual criteria:

\begin{itemize}
    \item \textbf{Overall Quality.} The participant judged the result based on the first overall visual impression, considering whether the image appeared clear, stable, plausible, and whether artifacts affected the viewing experience.
    \item \textbf{Surface Completeness.} The participant assessed whether the visible geometric surfaces were continuous and complete. For scene images, this criterion focused on obvious holes, discontinuities, or missing regions. For object images, it focused on missing, broken, or implausibly incomplete object surfaces.
    \item \textbf{Boundary Cleanliness.} The participant assessed whether boundaries between structures were clear and free from visible contamination. For scene images, this criterion focused on boundaries between objects and between objects and the background. For object images, it focused on whether the object's silhouette and outer boundary were clean and stable.
    \item \textbf{Naturalness.} The participant assessed whether the result was consistent with real visual experience. For scene images, this criterion focused on whether reconstructed or inserted regions were coherent with the surrounding scene in appearance, scale, and style. For object images, it focused on whether the object's shape and appearance resembled a normal, naturally occurring object.
\end{itemize}

All criteria were scored using a five-point scale following the Mean Opinion Score (MOS) protocol commonly used in subjective perceptual quality assessment. A score of 5 indicated excellent quality with almost no visible distortion; 4 indicated good quality with only minor distortion; 3 indicated fair quality with noticeable but still acceptable distortion; 2 indicated poor quality with severe distortion; and 1 indicated bad quality that was difficult to accept. The questionnaire instructions and score definitions were provided in both Chinese and English, and participants could choose either language before starting the study.

Because object-level renderings may occupy only a small region of the image canvas, the interface provided a local zoom and panning function for object questions. This function allowed participants to inspect object boundaries and local surface details more reliably. The zoom function did not reveal any hidden metadata and was not used for scene-level questions.

For each submitted questionnaire, the backend recorded the anonymous question identifier, hidden image type, hidden method label, hidden scene/object identifier, and the four criterion scores. After collection, we computed average scores and sample counts for each scene-method pair, each object-method pair, all scene images grouped by method, and all object images grouped by method. Aggregation was performed separately for each criterion and for the mean score across the four criteria.

\subsection{Auxiliary Agent Evaluation}
\label{sec:supp_agent_eval}

In addition to the human user study, we used \textbf{GPT-5.5} for an auxiliary assessment of scene and object quality. The evaluation followed the same four criteria and five-point scale as the human study: Overall Quality, Surface Completeness, Boundary Cleanliness, and Naturalness. Human ratings remained the primary measure of subjective quality.

We evaluated 50 matched scene pairs and 50 matched object pairs from the same six scenes in the Marble dataset. Each scene pair contained renderings of the original and converted scenes from the same viewpoint, while each object pair contained renderings of the corresponding original and regenerated object. This pairing allowed us to compare scores before and after conversion at both levels.

GPT-5.5 evaluated each image individually, returning four scores and brief justifications based on visible evidence, such as missing surfaces, blurred boundaries, detached fragments, or inconsistent appearance. We aggregated these scores separately from the human ratings.

The exact prompt used for the agent evaluation was:

\begin{quote}
\setlength{\parskip}{0.3em}
\setlength{\parsep}{0pt}
\setlength{\itemsep}{0pt}
\small
You are evaluating rendered 3D reconstruction images for a paper's supplementary qualitative audit. For each provided image, assign four independent MOS-style scores from 1 to 5. Use the following scale: 5 = excellent, almost no visible distortion; 4 = good, minor distortion that does not affect the overall viewing experience; 3 = fair, visible distortion but still acceptable; 2 = poor, severe distortion and degraded viewing experience; 1 = bad, very low quality and difficult to accept.

Evaluate the following four criteria independently:

1. Overall Quality: judge the first overall visual impression. Consider whether the rendering appears clear, stable, visually plausible, and whether artifacts affect the overall viewing experience. This criterion is holistic and should not be limited to one specific artifact type.

2. Surface Completeness: evaluate whether visible geometric surfaces are continuous and complete. For scene images, focus on holes, discontinuities, missing regions, or broken scene structures. For object images, focus on whether the object surface is missing, broken, collapsed, or implausibly incomplete.

3. Boundary Cleanliness: evaluate whether structural boundaries are clean and well separated. For scene images, focus on boundaries between objects and between objects and the background, including bleeding, halos, smearing, or boundary contamination. For object images, focus on the object's silhouette and outer contour, including edge blur, spillover, detached fragments, or unstable outlines.

4. Naturalness: evaluate whether the result is consistent with real visual experience. For scene images, focus on whether reconstructed or inserted regions are coherent with the surrounding scene in appearance, scale, lighting, and style. For object images, focus on whether the object's shape, material appearance, and proportions resemble a normal, naturally plausible object.

Do not infer or reveal method identity from sample identifiers. Base every score only on visible image evidence. For each image, output the category, sample identifier, four numeric scores, and a concise rationale that explicitly mentions the main evidence for the assigned scores.
\end{quote}

\section{More results}
\label{app:more_results}

To further demonstrate the robustness and generalization of our pipeline, we include additional qualitative results in the supplementary material. Figures~\ref{fig:more1} and~\ref{fig:more2} show extended visualizations on the datasets, illustrating the full pipeline progression from the input scene to segmentation, removal, inpainting, and final assembly.

\newpage
\begin{figure}[H]
    \centering
    \includegraphics[width=\linewidth]{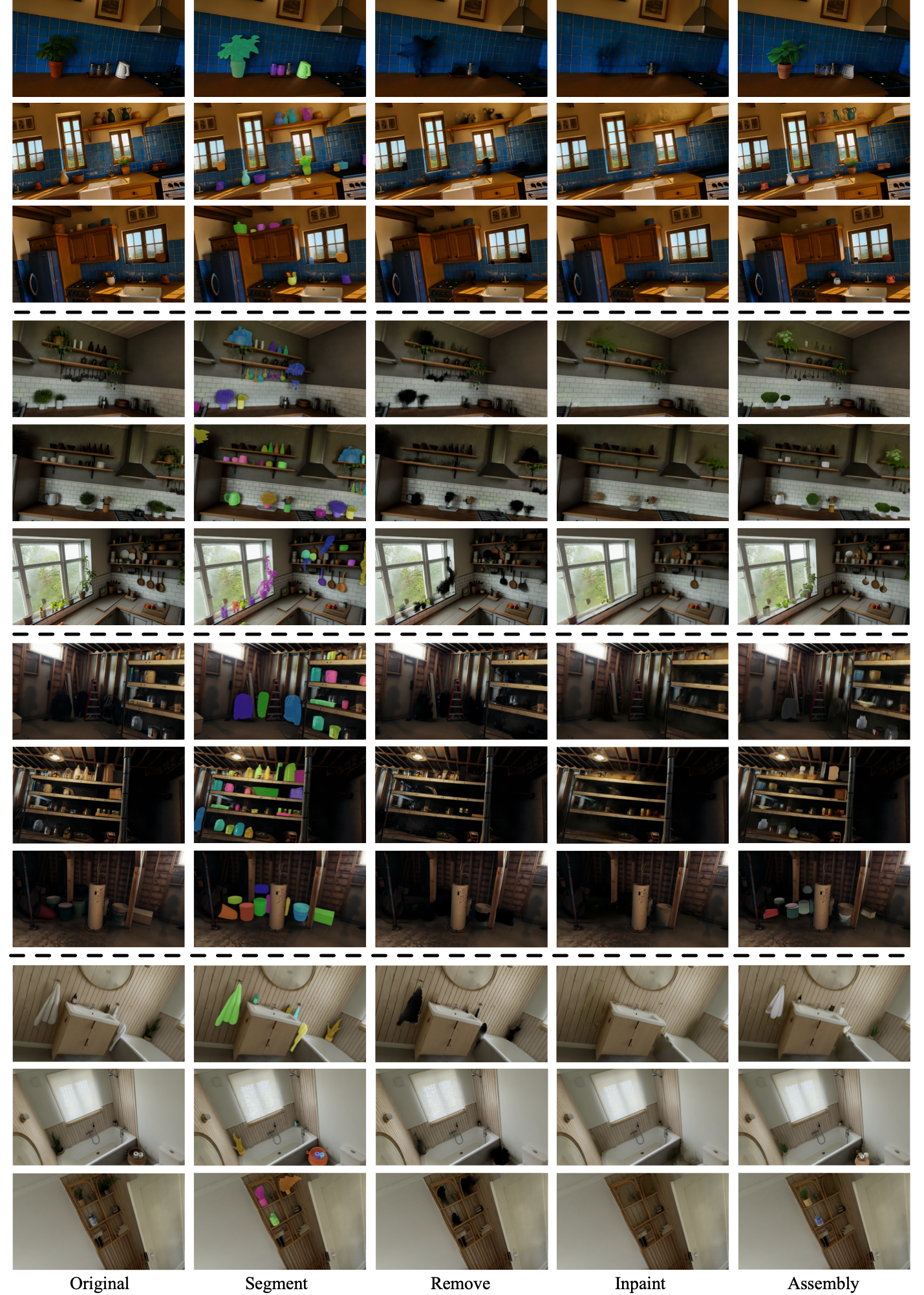}
    
    \caption{Additional pipeline results. For each scene, we show from left to right: the original Marble-generated 3DGS scene with detected object masks, object removal, background inpainting, and final assembly with generated objects placed back.}
    \label{fig:more1}
\end{figure}

\begin{figure}[H]
    \centering
    \includegraphics[width=\linewidth]{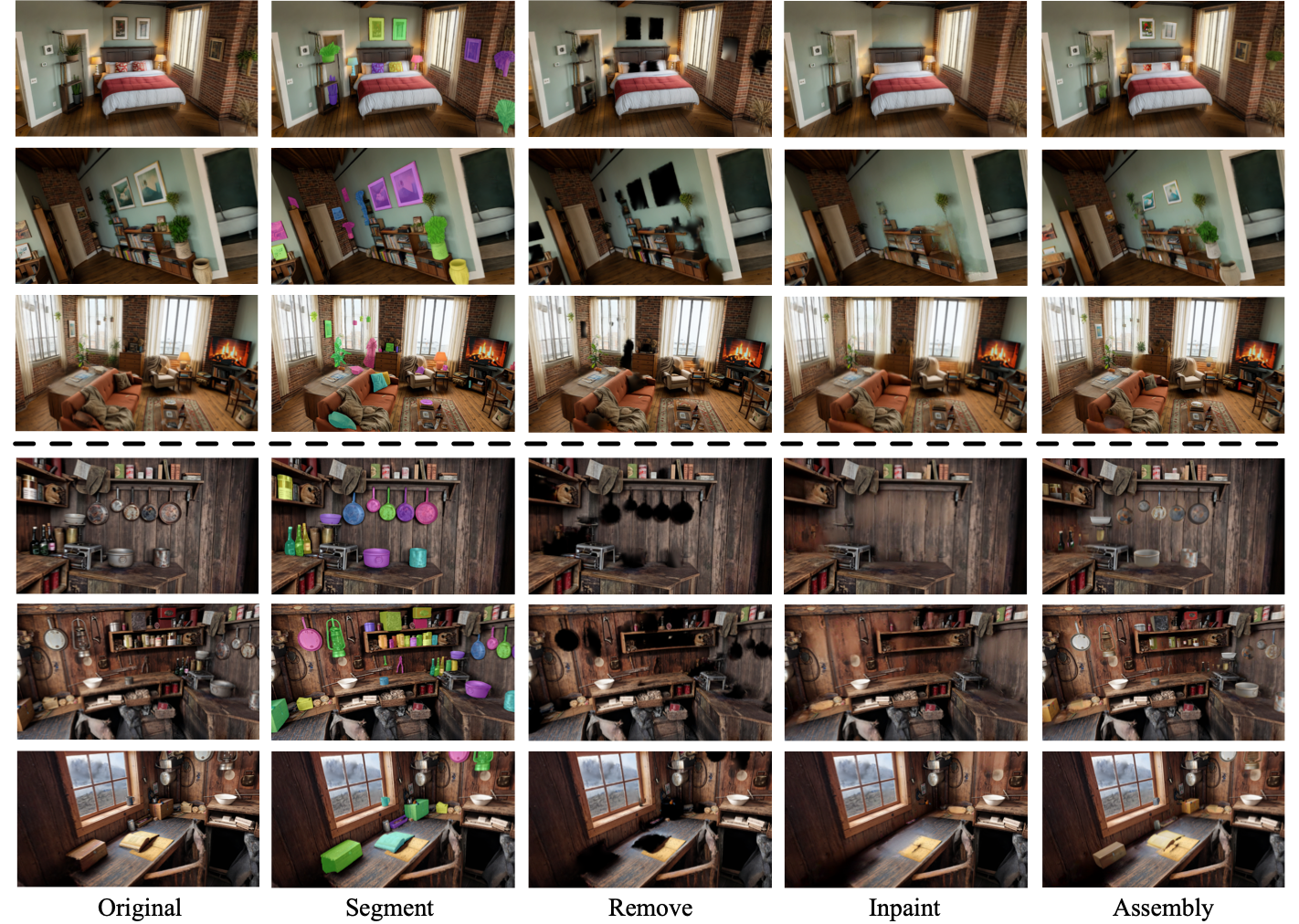}
    
    \caption{Additional pipeline results, which contains highly cluttered and occluded scenes. Despite the increased complexity, our method successfully decomposes objects, repairs the background, and reassembles high-quality assets.}
    \label{fig:more2}
\end{figure}

\end{document}